\documentclass[letterpaper, 10 pt, conference]{ieeeconf}
\IEEEoverridecommandlockouts
\usepackage{amsmath,amssymb,amsfonts}
\usepackage{graphicx}
\usepackage{textcomp}

\usepackage{xcolor}
\usepackage[normalem]{ulem}

\def\BibTeX{{\rm B\kern-.05em{\sc i\kern-.025em b}\kern-.08em
    T\kern-.1667em\lower.7ex\hbox{E}\kern-.125emX}}
    
\bstctlcite{IEEEexample:BSTcontrol}

\title{Comparing Feedback Linearization and Adaptive Backstepping Control for Airborne Orientation of Agile Ground Robots using Wheel Reaction Torque}

\author{Jinho Kim$^{1}$, \IEEEmembership{Member,~IEEE}, Daniel J. Gonzalez$^{1}$, \IEEEmembership{Member,~IEEE},\\ 
	and Christopher M. Korpela$^{1}$, \IEEEmembership{Senior Member,~IEEE}
	\thanks{$^{1}$Robotics Research Center, Department of Electrical Engineering and Computer Science. United States Military Academy, West Point, NY 10996, USA. Email: {\tt\small \{jinho.kim, daniel.gonzalez, christopher.korpela\}@westpoint.edu}.}
}
    
\begin{document}
\maketitle

\begin{abstract}\label{ABS}
In this paper, two nonlinear methods for stabilizing the orientation of a Four-Wheel Independent Drive and Steering (4WIDS) robot while in the air are analyzed, implemented in simulation, and compared. AGRO (the Agile Ground Robot) is a 4WIDS inspection robot that can be deployed into unsafe environments by being thrown, and can use the reaction torque from its four wheels to command its orientation while in the air. Prior work has demonstrated on a hardware prototype that simple PD control with hand-tuned gains is sufficient, but hardly optimal, to stabilize the orientation in under 500ms. 

The goal of this work is to decrease the stabilization time and reject disturbances using nonlinear control methods. A model-based Feedback Linearization (FL) was added to compensate for the nonlinear Coriolis terms. However, with external disturbances, model uncertainty and sensor noise, the FL controller does not guarantee stability. As an alternative, a second controller was developed using backstepping methods with an adaptive compensator for external disturbances, model uncertainty, and sensor offset. The controller was designed using Lyapunov analysis. A simulation was written using the full nonlinear dynamics of AGRO in an isotropic steering configuration in which control authority over its pitch and roll are equalized. The PD+FL control method was compared to the backstepping control method using the same initial conditions in simulation. Both the backstepping controller and the PD+FL controller stabilized the system within 250 milliseconds. The adaptive backstepping controller was also able to achieve this performance with the adaptation law enabled and compensating for offset noisy sinusoidal disturbances.

\end{abstract}

\section{Introduction}\label{SEC:INT}
The unmanned ground vehicle (UGV) has been the subject of much research and application toward numerous missions and goals that present a danger to human first responders such as exploration, reconnaissance, and search and rescue operations \cite{Delmerico2017,Perez-Imaz2016,Tadokoro2019}. The most widely used wheeled ground robots have skid-steer \cite{Yi2009} or Ackerman \cite{Weinstein2010} steering designs, so their movement is limited compared to legged robots that can move omnidirectionally. In order to maximize the capability of wheeled UGVs, different steering geometries have been considered.

\begin{figure}[t]
	\centering
	\includegraphics[width=1\linewidth]{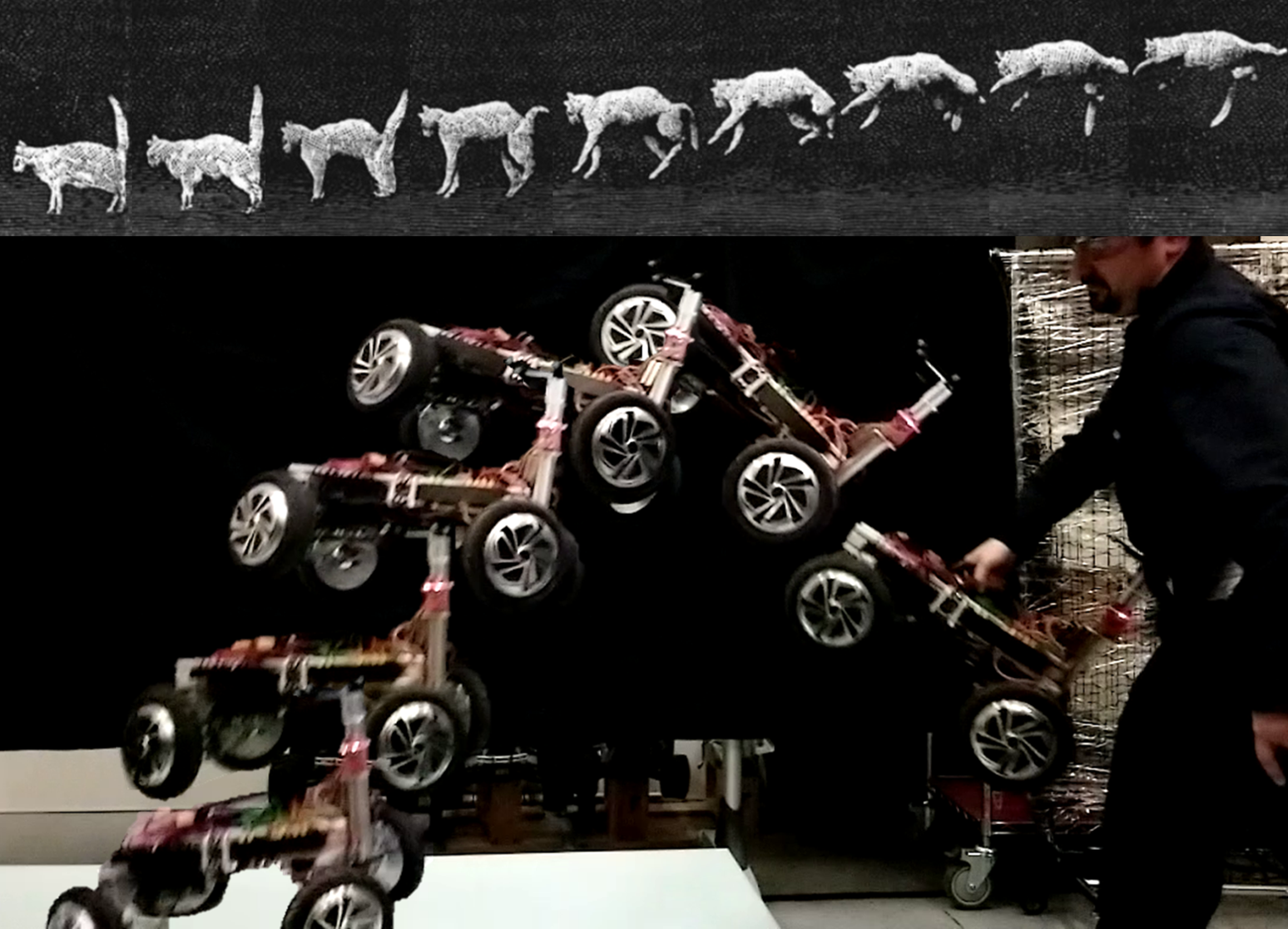}
	\caption{Composite video stills of (top) a cat orienting its body mid-air and landing on its feet \cite{Marey1894} and (bottom) AGRO being thrown, stabilizing its attitude mid-air, and landing on all four wheels.}
	\label{fig01}
\end{figure}

AGRO is a novel Agile Ground RObot that aims to be highly maneuverable and rapidly deployable with the best attributes of both wheeled and legged robots. AGRO has the ability to maneuver quasi-omnidirectionally on the ground with a Four-Wheel Independent Drive and Steering (4WIDS) architecture similar to \cite{Mori2002} and \cite{Michaud2003}. This wheel architecture also allows AGRO the novel ability to control its orientation in the air by using the reaction torques from in-wheel hub motors to ensure it lands upright \cite{Gonzalez2020}. Inspired by the agile mobility of a cat, this capability allows AGRO to evenly distribute the force of impact to all four wheels when it is thrown over walls and fences, or through windows to achieve rapid and reliable deployment (See Fig. \ref{fig01}).

While airborne, AGRO exhibits complicated dynamics with coupled states. Since simple linear control design techniques can be easily employed, several linearization approaches have been proposed to control similar nonlinear systems \cite{Cambera2017,Buss2016,Farzan2019,Terry2017,Engelhardt2016}. One of the most commonly used control designs is a feedback linearization method that algebraically transforms the nonlinear system dynamics into a fully or partially linearized system by choosing the control inputs to cancel the nonlinear terms. In order to achieve the best results, however, exact knowledge of the system parameters is required, and these are difficult to fully model or acquire empirically.

Another applicable nonlinear control strategy is the backstepping control technique that was introduced in 1992 \cite{kokotovic1992joy}. Backstepping control design shows more flexibility compared to the feedback linearization because the resulting input-output dynamics need not be linear. Additionally, backstepping controllers keep using useful nonlinear terms whereas a typical feedback linearization approach cancels nonlinear terms. The main idea of the backstepping control strategy is to use some of the state variables as virtual- or pseudo-controls, and depending on the dynamics of each state, design intermediate control laws that are verified using Lyapunov analysis.

This paper presents two nonlinear controllers for the AGRO system. First, we apply a feedback linearization approach to control $\varphi$-$\theta$-$\psi$. To apply the feedback linearization controller without complicated calculation, we simplify the equation of system dynamics by defining specific control input terms. Although the feedback linearization controller is simple to implement, there is the possibility that the model uncertainty and external disturbances can cause instability of the system or performance degradation because it uses inverse system dynamics as part of the control input to cancel nonlinear terms. To consider the robustness issue with model uncertainty and external disturbances, we apply the adaptive backstepping control strategy for controlling AGRO using a virtual control input.

In this paper, the dynamics of an airborne 4WIDS robot are derived in Section II. In Section III, the feedback linearization strategy for stabilizing aerial attitude is described. Then the adaptive backstepping control approach is presented with system uncertainties and unknown disturbances in Section IV. In Section V, the simulation results of the two controllers are presented. Section VI provides a conclusion and outlines future work to be conducted.

\section{Dynamics of an Airborne 4WIDS Robot}
\begin{figure}[t]
	\centering
	\includegraphics[width=0.8\linewidth]{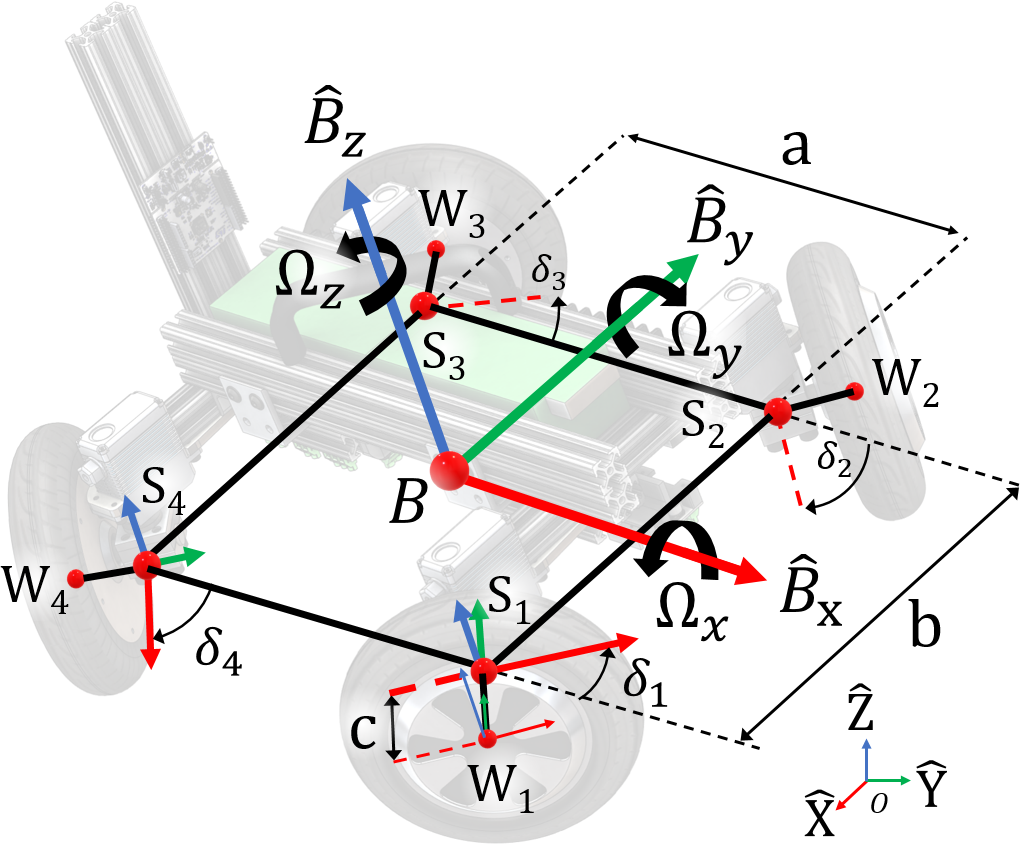}
	\caption{3D Airborne Kinematic Parameters. Steering angle displacement $\delta_i$ is measured deviation from the forward driving direction, with $\delta_i = \angle \hat{B}_X \hat{S}_{xi}$ for $i$ = 1, 4 and $\delta_i = \angle\hat{B}_X \hat{S}_{xi}-\pi$ for $i$ = 2, 3. Unit Vectors $\hat{B}_X$, $\hat{B}_Y$, and $\hat{B}_Z$ are fixed in the Newtonian frame $O$}
	\label{Kinematic Params 3D}
\end{figure}

To design a controller for the orientation of AGRO while in the air, the dynamic model of the system must be derived. The following is a summarized derivation of the airborne dynamics of AGRO. For further understanding of kinematics parametrization, singularity analysis, and its affect on aerial orientation manipulability, as well as a more complete derivation of the following dynamics, please refer to \cite{Gonzalez2020}.

\begin{figure}[t]
	\centering
	\includegraphics[width=1\linewidth]{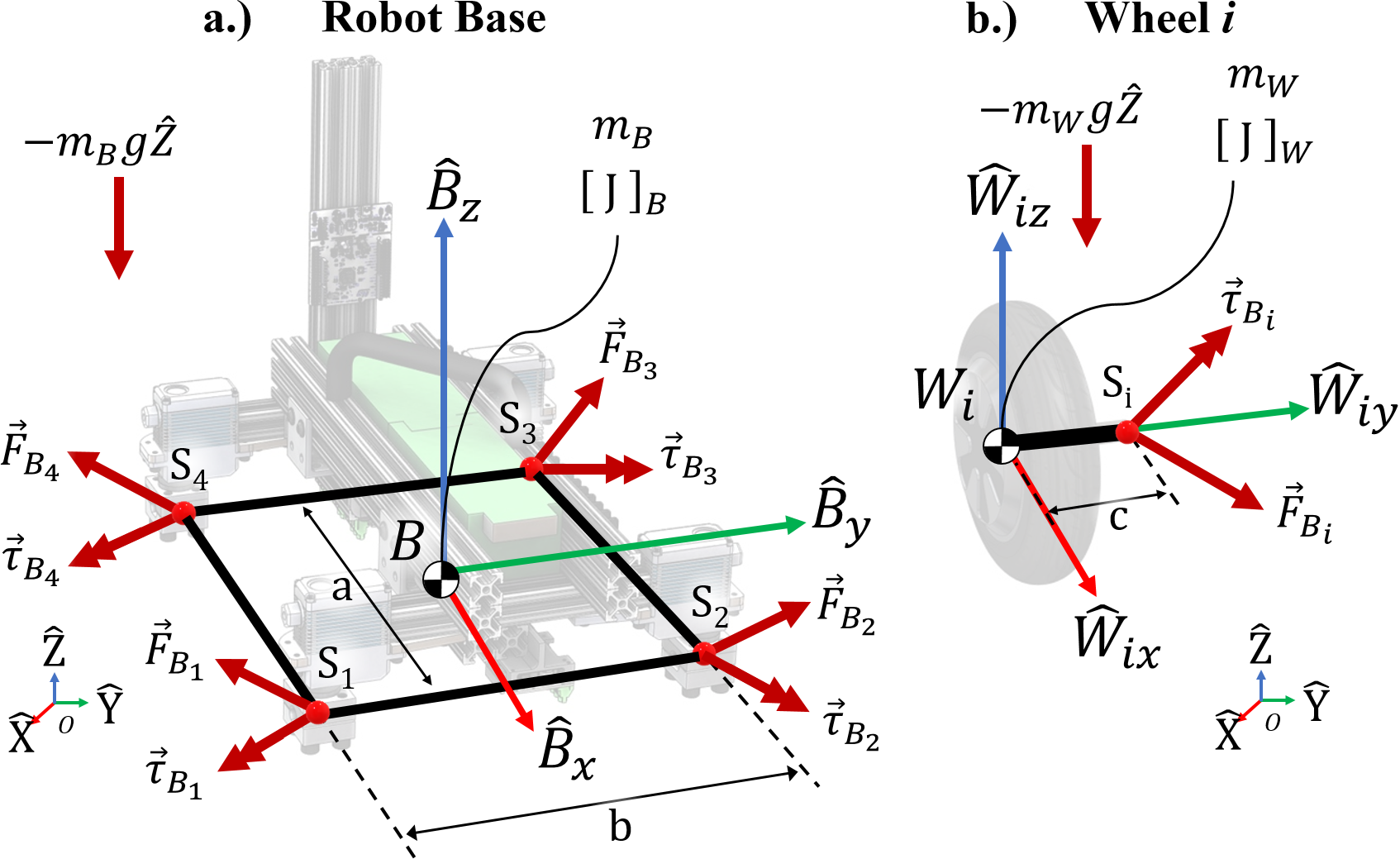}
	\caption{Free Body Diagram for a) the main base and b) a single wheel module.}
	\label{FBD}
\end{figure}

The four attached steerable offset wheels impart reaction torques $\Vec{\tau}_{Bi}$ and forces $\Vec{F}_{Bi}$ on the base at points $S_i$ (See Fig. \ref{FBD}). Taking the Newton-Euler equation for angular momentum for the base about its mass center, we get
\begin{equation}
    \sum_{i=1}^{4}\left(\Vec{\tau}_{Bi}+\Vec{r}_{BS_i}\times\Vec{F}_{Bi}\right) =
    \frac{d}{dt}\left(\left[J_B\right]\Vec{\Omega}\right)
\end{equation}

Torque from each wheel $\Vec{\tau}_{Bi}$ can be broken down as the commanded wheel drive torque input $\tau_i$, the commanded steering joint torque $\tau_{\delta i}$ and one reaction torque $\tau_x$ about the wheel’s local X axis. 
\begin{equation}\label{torques}
    \Vec{\tau}_{Bi} = 
    \begin{cases}
    \tau_{xi} \hat{W_i}_x \\
    \tau_i \hat{W_i}_y \\
    \tau_{\delta i} \hat{W_i}_z    
    \end{cases}
\end{equation}

This scalar wheel reaction torque component $\tau_{xi}$ can be derived by dotting the total angular momentum of the wheel about the steering joint by its local $\hat{W}_{ix}$ axis.
\begin{equation}\label{Tauxi}
    \tau_{xi}=\left(-\frac{d}{dt}\left(\left[J_W\right]\Vec{\omega}_i\right)-\Vec{r}_{W_iS_i}\times\Vec{F}_{Bi}\right) \cdot \hat{W}_{ix}.
\end{equation}

Each wheel has linear momentum, is acted upon by gravity, and reacts against the robot base. Resultant reaction forces $\Vec{F}_{Bi}$ can be solved for using the linear momentum equation for the wheels.
\begin{equation}\label{FBI}
    \Vec{F}_{Bi} =
    -m_W\left(\frac{d^2}{dt^2}\left(\Vec{r}_{OW_i}\right)+g\hat{Z}\right)
\end{equation}
For tractability, we assume nonmoving steering angles ($\dot\delta=0$, $\ddot\delta=0$), cross-symmetry ($\delta_1 = \delta_3$, $\delta_2 = \delta_4$), and cross-symmetric torque application ($\tau_1 = -\tau_3$, $\tau_2 = -\tau_4$).

Combining these expressions leads to the three nonlinear equations of motion for AGRO’s orientation while in the air.
\begin{equation} \label{EOMx}
\begin{split}
(J_{Bxx}+J_{m_Wxx}+2J_{Wxx}(\cos\delta_1+\cos\delta_2))\dot{\Omega}_x&\\
+(J_{Bzz}-J_{Byy} + J_{m_Wxx})\Omega_y\Omega_z&\\
= 2\tau_1 \sin\delta_1-2\tau_2\sin&\delta_2
\end{split}
\end{equation}
\begin{equation}\label{EOMy}
\begin{split}
(J_{Byy}+J_{m_Wyy}+2J_{Wxx}(\sin\delta_1+\sin\delta_2))\dot\Omega_y&\\
+(J_{Bxx}-J_{Bzz}+ J_{m_Wyy})\Omega_x\Omega_z&\\
=-2\tau_1 \cos \delta_1+2\tau_2\cos&\delta_2
\end{split}
\end{equation}
\begin{equation} \label{EOMz}
\begin{split}
(J_{Bzz})\dot\Omega_z + (J_{Byy}-J_{Bxx})\Omega_x\Omega_y&=4\tau_{\delta}
\end{split}
\end{equation}
where
\begin{equation} \label{JmWxx}
    J_{m_Wxx} = 2m_W\left((\frac{b}{2} + c\cos\delta_1)^2+(\frac{b}{2} + c\cos\delta_2)^2\right)
\end{equation}
and 
\begin{equation} \label{JmWyy}
    J_{m_Wyy} = 2m_W\left((\frac{a}{2} + c\sin\delta_1)^2+(\frac{b}{2} + c\sin\delta_2)^2\right)
\end{equation}
In \eqref{EOMx}-\eqref{JmWyy}, $J_{Bii}$ and $J_{Wii}$ are the body and wheel moment of inertia along $\hat{B}_{i}$ and $\hat{W}_{i}$, respectively. $J_{m_{W}ii}$ is the overall inertia of the base reflected by wheel masses. $\delta_i$ is steering angle displacement measured deviation from the forward driving direction. Each equation of motion \eqref{EOMx}-\eqref{EOMz} contains linear inertial terms, nonlinear Coriolis terms, and control inputs from wheel drive and steering torques.

The input commands can be further simplified using the Jacobian relating whole body reaction torques about the body-centric axes to the wheel and steering torques: 
\begin{equation}\label{torqueMatrix}
    \begin{bmatrix}\tau_x\\ \tau_y \\ \tau_z\end{bmatrix} = 
    \mathbb{J}_\tau\begin{bmatrix}\tau_1\\ \tau_2 \\ \tau_\delta\end{bmatrix}
\end{equation}
where
\begin{equation}\label{Jacobian}
    \mathbb{J}_\tau= 
    \begin{bmatrix}
    2\sin(\delta_1) & -2\sin(\delta_2) & 0\\
    -2\cos(\delta_1) & 2\cos(\delta_2) & 0\\
    0 & 0 & 4
    \end{bmatrix}
\end{equation}

\section{Feedback Linearization Control}\label{SEC:FL}
The ability to reliably and repeatably land on its wheels is critical for the continuation of AGRO's inspection and response missions. This section proposes an amendment to the original simple PD control strategy by using feedback linearization to enable AGRO to compensate for nonlinear dynamics and land more reliably.

First, the equations of motion \eqref{EOMx}-\eqref{EOMz} can be rewritten with [$\Omega_x$, $\Omega_y$, $\Omega_z$] = [$\dot\varphi$, $\dot\theta$, $\dot\psi$] as
\begin{equation} \label{eq_EOM1}
\begin{split}
    &\ddot{\varphi} =\frac{(J_{Byy}-J_{Bzz}-J_{m_{W}xx})\dot{\theta}\dot{\psi}+\tau_x}{J_{Bxx}+J_{m_{W}xx}+2J_{Wxx}(cos\delta_{1}+cos\delta_{2})}\\
\end{split}
\end{equation}
\begin{equation} \label{eq_EOM2}
\begin{split}
    &\ddot{\theta} = \frac{(-J_{Bxx}+J_{Bzz}-J_{m_{W}yy})\dot{\varphi}\dot{\psi}+\tau_y}{J_{Byy}+J_{m_{W}yy}+2J_{Wxx}(sin\delta_{1}+sin\delta_{2})}
\end{split}
\end{equation}
\begin{equation} \label{eq_EOM3}
\begin{split}
    \ddot{\psi} = \frac{1}{J_{Bzz}}\left((J_{Bxx}-J_{Byy})\dot{\varphi}\dot{\theta}+\tau_z\right)
\end{split}
\end{equation}
where $[\varphi, \theta, \psi]$ represents the small rotation roll, pitch, and yaw angles, respectively.

We define input terms as
\begin{equation} \label{eq_inputs}
u_1=\tau_x,~~ u_2=\tau_y,~~ u_3=\tau_z
\end{equation}
and simplify inertial constants as
\begin{equation}
    \begin{split}    
J_{\varphi,1} &= \left(J_{Bxx}+J_{m_{W}xx}+2J_{Wxx}(cos\delta_1+cos\delta_2)\right)\\
J_{\theta,1} &= \left(J_{Byy}+J_{m_{W}yy}+2J_{Wxx}(sin\delta_1+sin\delta_2)\right)\\
J_{\psi,1} &= J_{Bzz}\\
J_{\varphi,2}&=J_{Byy}-J_{Bzz}-J_{m_{W}xx}\\
J_{\theta,2}&=-J_{Bxx}+J_{Bzz}-J_{m_{W}yy}\\
J_{\psi,2}&=J_{Bxx}-J_{Byy}
    \end{split}
\end{equation}
to simplify the equations of motion of the AGRO system to
\begin{equation} \label{eq_EOM_phi}
\ddot{\varphi}= \frac{J_{\varphi,2}\dot{\theta}\dot{\psi} + u_1}{J_{\varphi,1}}
\end{equation}
\begin{equation} \label{eq_EOM_theta}
\ddot{\theta}= \frac{J_{\theta,2}\dot{\varphi}\dot{\psi} + u_2}{J_{\theta,1}}
\end{equation}
\begin{equation} \label{eq_EOM_psi}
\ddot{\psi}= \frac{J_{\psi,2}\dot{\varphi}\dot{\theta} + u_3}{J_{\psi,1}},
\end{equation}

For a vector form of the system, (\ref{eq_EOM_phi})-(\ref{eq_EOM_psi}) can be written with $\textbf{x} = [\varphi, \theta, \psi]^T$ and $\textbf{u}=[u_1, u_2, u_3]^T$ as the following
\begin{equation} \label{eq_EOM_matrix}
\begin{split}
\ddot{\textbf{x}} &= f(\textbf{x},\dot{\textbf{x}})+g(\textbf{x})\textbf{u}\\
f(\textbf{x},\dot{\textbf{x}}) =
\begin{bmatrix}
\frac{J_{\varphi,2}}{J_{\varphi,1}}\dot{\theta}\dot{\psi}\\
\frac{J_{\theta,2}}{J_{\theta,1}}\dot{\varphi}\dot{\psi}\\
\frac{J_{\psi,2}}{J_{\psi,1}}\dot{\varphi}\dot{\theta}\\
\end{bmatrix}&, \quad
g(\textbf{x}) =
\begin{bmatrix}
\frac{1}{J_{\varphi,1}} & 0 & 0\\
0 & \frac{1}{J_{\theta,1}} & 0\\
0 & 0 & \frac{1}{J_{\psi,1}}
\end{bmatrix}
\end{split}
\end{equation}

\begin{figure}[t]
	\centering
	\includegraphics[width=1\linewidth]{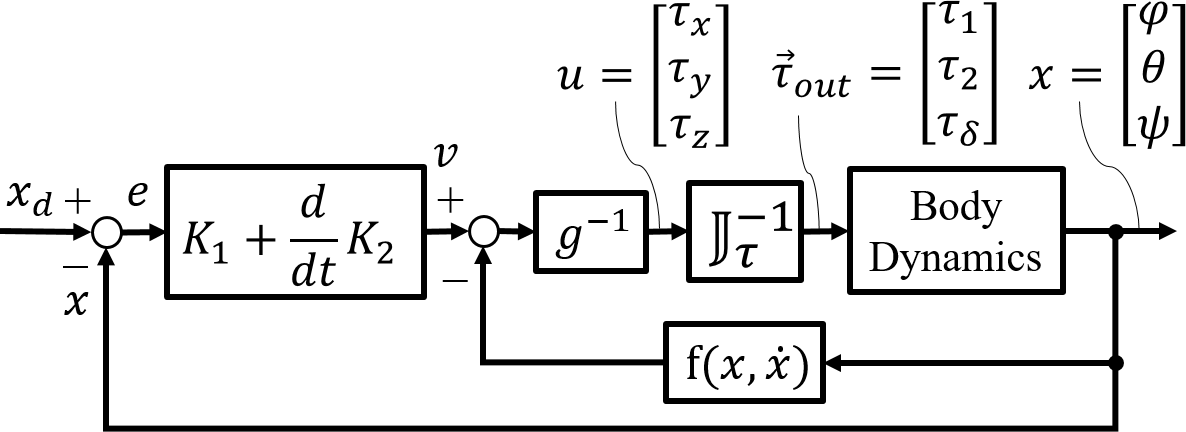}
	\caption{Block diagram of the PD + Feedback Linearization control architecture.}
	\label{figFLBlock}
\end{figure}
From (\ref{eq_EOM_matrix}), we apply the feedback linearization method for $\varphi$-$\theta$-$\psi$ control of the AGRO system, as shown in Fig. \ref{figFLBlock}. To obtain control inputs for the $\varphi$-$\theta$-$\psi$ controller, we choose
\begin{equation} \label{eq_FL_u}
\begin{bmatrix}
    u_1\\
    u_2\\
    u_3
\end{bmatrix}
=
g(x)^{-1}
\begin{bmatrix}
    -\frac{J_{\varphi,2}}{J_{\varphi,1}}\dot{\theta}\dot{\psi}+v_1\\
    -\frac{J_{\theta,2}}{J_{\theta,1}}\dot{\varphi}\dot{\psi}+v_2\\
    -\frac{J_{\psi,2}}{J_{\psi,1}}\dot{\varphi}\dot{\theta}+v_3\\
\end{bmatrix}
\end{equation}
where [$v_1$, $v_2$, $v_3$] are pseudo-inputs. Setting pseudo input terms as shown in the following
\begin{equation} \label{eq_FL_pseudo_u}
\begin{split}
    v_1 = \ddot{\varphi}_d + k_{\varphi,1}\dot{e}_{\varphi} + k_{\varphi,2}e_{\varphi}\\
    v_2 = \ddot{\theta}_d + k_{\theta,1}\dot{e}_{\theta} + k_{\theta,2}e_{\theta}\\
    v_3 = \ddot{\psi}_d + k_{\psi,1}\dot{e}_{\psi} + k_{\psi,2}e_{\psi}
\end{split}
\end{equation}
yields
\begin{equation}
\begin{split}
    \ddot{e}_{\varphi} + k_{\varphi,1}\dot{e}_{\varphi} + k_{\varphi,2}e_{\varphi} = 0\\
    \ddot{e}_{\theta} + k_{\theta,1}\dot{e}_{\theta} + k_{\theta,2}e_{\theta} = 0\\
    \ddot{e}_{\psi} + k_{\psi,1}\dot{e}_{\psi} + k_{\psi,2}e_{\psi} = 0
\end{split}
\end{equation}
where $e_{\varphi}=\varphi_d-\varphi$, $e_{\theta}=\theta_d-\theta$, $e_{\psi}=\psi_d-\psi$ and $k_{i,j}$ are the diagonal components of gain matrices $K_j$. By using these control inputs, the nonlinear terms in the system are canceled, and the system can be stable following a given reference input.

\section{Adaptive Backstepping Control}
We now consider an adaptive backstepping method of designing the feedback controller for AGRO, which can more readily counteract unmodeled external disturbances or modeling error.

Considering model uncertainties and external disturbance, the AGRO system (\ref{eq_EOM_matrix}) can be rewritten as
\begin{equation} \label{eq_dyn_with_L}
\begin{split}
    \ddot{\textbf{x}} &= [f(\textbf{x},\dot{\textbf{x}}) + \Delta f(\textbf{x},\dot{\textbf{x}})]  + [g(\textbf{x}) + \Delta g(\textbf{x})]\textbf{u} + \varepsilon \\
    &= f(\textbf{x},\dot{\textbf{x}}) + g(\textbf{x})\textbf{u} + L,
\end{split}
\end{equation}
where 
\begin{equation}
L = \Delta f(\textbf{x},\dot{\textbf{x}}) + \Delta g(\textbf{x}) \textbf{u} + \varepsilon
\end{equation}
where $\Delta f(\textbf{x},\dot{\textbf{x}})$ and $\Delta g(\textbf{x})$ are the unknown uncertainties, $\varepsilon$ represents an unknown external disturbance.

The control objective is to force $\textbf{x}$ to track a given reference signal $\textbf{x}_d$. \textbf{Specifically, our control objective is as follows:} Given the desired state variables, $\textbf{x}_d$, determine a backstepping controller, so the output errors are as small as possible under the constraints. Here, we assume that the given desired state variables are bounded as follows
\begin{equation}
    \vert \textbf{x}_d \vert ^2 + \vert \dot{\textbf{x}}_d \vert ^2 + \vert \ddot{\textbf{x}}_d \vert ^2 \leq \rho
\end{equation}
where $\rho$ is a positive constant.

Let $\textbf{e}_1 = \textbf{x}_d - \textbf{x}$ define the error vector with respect to the vector of desired state variables $\textbf{x}_d = [\varphi_d, \theta_d, \psi_d]^T$, and let the first Lyapunov function be
\begin{equation} \label{eq_V1}
    V_1 = \frac{1}{2} \textbf{e}_1^T \textbf{e}_1.
\end{equation}
Then we have the derivative of $V_1$ as
\begin{equation}
\begin{split} \label{eq_V1-2}
    \dot{V}_1 &=  \textbf{e}_1^T \dot{\textbf{e}}_1 \\
    &= \textbf{e}_1^T (\dot{\textbf{x}}_d - \dot{\textbf{x}}).
\end{split}
\end{equation}
Here, we set $\dot{\textbf{x}}$ as a virtual control and define the desired value of virtual control, known as a stabilizing function, as follows
\begin{equation} \label{eq_U1}
    U_v = \dot{\textbf{x}}_d + K_1\textbf{e}_1,
\end{equation}
where $K_1 = diag[k_{11},k_{12},k_{13}]$ is a diagonal gain matrix with positive entries.
Substituting $\dot{\textbf{x}}$ in (\ref{eq_V1-2}) with (\ref{eq_U1}), the derivative of $V_1$ becomes
\begin{equation}
    \dot{V}_1 = -K_1 \textbf{e}_1^T \textbf{e}_1 \leq 0.
\end{equation}

Now, the deviation of the virtual control from its desired value can be defined as
\begin{equation} \label{eq_e2}
    \textbf{e}_2 = U_v - \dot{\textbf{x}} = \dot{\textbf{x}}_d - \dot{\textbf{x}} + K_1\textbf{e}_1.
\end{equation}
The derivative of $\textbf{e}_2$ can be presented as
\begin{equation} \label{eq_e2_dot}
\begin{split}
    \dot{\textbf{e}}_2 &= \ddot{\textbf{x}}_d  - \ddot{\textbf{x}} + K_1\dot{\textbf{e}}_1 \\
    &=\ddot{\textbf{x}}_d -  f(\textbf{x},\dot{\textbf{x}}) - g(\textbf{x})\textbf{u} - L + K_1\dot{\textbf{e}}_1.
\end{split}
\end{equation}

Let us define $\hat{L}$ as the estimated value of $L$ and $\Tilde{L}:=L-\hat{L}$, and let the second Lyapunov function be
\begin{equation} \label{eq_V2}
    V_2 = \frac{1}{2} \textbf{e}_1^T \Gamma \textbf{e}_1 + \frac{1}{2} \textbf{e}_2^T \Lambda \textbf{e}_2 + \frac{1}{2}\Tilde{L}^T \Sigma\Tilde{L},
\end{equation}
where $\Gamma$, $\Lambda$, and $\Sigma$ are positive semi-definite weighting matrices.
Assuming that $L$ changes slowly enough, which leads to $\frac{d}{dt}\Tilde{L}=\dot{\Tilde{L}} \approx -\dot{\hat{L}}$, the first-order derivative of $V_2$ can be derived as
\begin{equation} \label{eq_V2-2_1}
    \dot{V}_2 = \textbf{e}_1^T \Gamma \dot{\textbf{e}}_1 + \textbf{e}_2^T \Lambda \dot{\textbf{e}}_2 + \Tilde{L}^T\Sigma(-\dot{\hat{L}})
\end{equation}
Then \eqref{eq_V2-2_1} becomes
\begin{equation} \label{eq_V2-2_2}
\begin{split}
    \dot{V}_2 &= \textbf{e}_1^T \Gamma (\dot{\textbf{x}}_d-\dot{\textbf{x}}) + \textbf{e}_2^T \Lambda (\ddot{\textbf{x}}_d - \ddot{\textbf{x}}   + K_1\dot{\textbf{e}}_1) - \Tilde{L}^T \Sigma \dot{\hat{L}}\\
    &= \textbf{e}_1^T \Gamma (\textbf{e}_2 - K_1 \textbf{e}_1) \\
    & \ \ + \textbf{e}_2^T \Lambda (\ddot{\textbf{x}}_d -  f(\textbf{x},\dot{\textbf{x}}) - g(\textbf{x})\textbf{u} - L + K_1\dot{\textbf{e}}_1) - \Tilde{L}^T \Sigma \dot{\hat{L}}\\
\end{split}
\end{equation}

To make the first-order derivative of $V_2$ negative definite, the backstepping control input should be selected as
\begin{equation} \label{eq_UB}
    U_{B} =g^{-1}(\textbf{x}) (\Lambda^{-1}\Gamma\textbf{e}_1 - f(\textbf{x},\dot{\textbf{x}}) - \hat{L} + \ddot{\textbf{x}}_d  + K_1\dot{\textbf{e}}_1 + K_2 \textbf{e}_2).
\end{equation}
Then \eqref{eq_V2-2_2} with \eqref{eq_UB} becomes
\begin{equation} \label{eq_V2_final}
\begin{split}
    \dot{V}_2 &= - K_1 \textbf{e}_1^T \Gamma \textbf{e}_1 - K_2 \textbf{e}_2^T \Lambda \textbf{e}_2 - \textbf{e}_2^T \Lambda (L - \hat{L}) - \Tilde{L}^T \Sigma \dot{\hat{L}} \\
    &= - K_1 \textbf{e}_1^T \Gamma \textbf{e}_1 - K_2 \textbf{e}_2^T \Lambda \textbf{e}_2 - \Tilde{L}^T(\Lambda \textbf{e}_2 + \Sigma \dot{\hat{L}})
\end{split}
\end{equation}
\begin{figure}[t]
	\centering
	\includegraphics[width=1\linewidth]{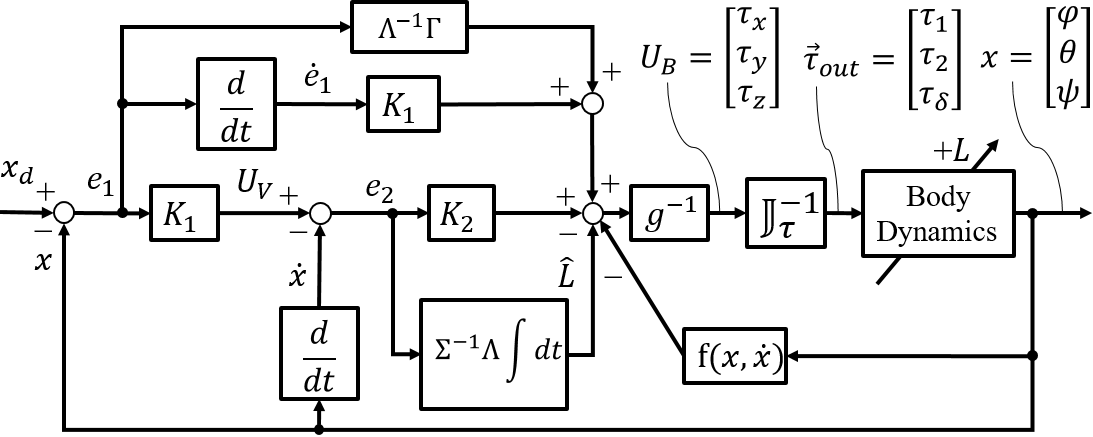}
	\caption{Block diagram of the Backstepping control architecture with $\dot{\textbf{x}_d}=0$ and $\ddot{\textbf{x}}_d=0$.}
	\label{figBSBlock}
\end{figure}

The term $K_2 \textbf{e}_2$ is added to stabilize the output error $\textbf{e}_1$.
With positive entries of $K_2 = diag[k_{21},k_{22},k_{23}]$ and an adaptive update rule
\begin{equation}
\dot{\hat{L}} = -\Sigma^{-1}\Lambda \textbf{e}_2
\end{equation}
we can write (\ref{eq_V2_final}) as
\begin{equation} \label{eq_V2_stability}
    \dot{V}_2 = - K_1 \textbf{e}_1^T \Gamma \textbf{e}_1 - K_2 \Lambda \textbf{e}_2^T \textbf{e}_2 \leq 0
\end{equation}
which means $e_1$ and $e_2$ go to zero and the system is stable. The estimation $\hat{L}$ from $\frac{d}{dt}{\hat{L}}$ will converge to $L$ in steady state, so $\dot{V}_2 $ is negative semidefinite. Thus the system achieves Globally Asymptotic Stability. 

\section{Simulation and Comparison}
In this section, we present simulation results of the two controllers; Proportional-Derivative with Feedback Linearization (PD+FL) and adaptive Backstepping (BS). The unknown uncertainties and external disturbances are not considered in the first comparison. The same simulation is then performed for the adaptive Backstepping controller with unknown uncertainties and external disturbances, to test the adaptation law. Note that we do not consider $\psi$ control in this simulation section, because the objective of these control strategies is to enable ARGO to land on the ground with stabilized attitude, $i.e.$ $\phi = 0\ deg.$ and $\theta = 0\ deg.$.

\subsection{Comparing Feedback Linearization and Backstepping}
\begin{figure}[t]
	\centering
	\includegraphics[width=1\linewidth]{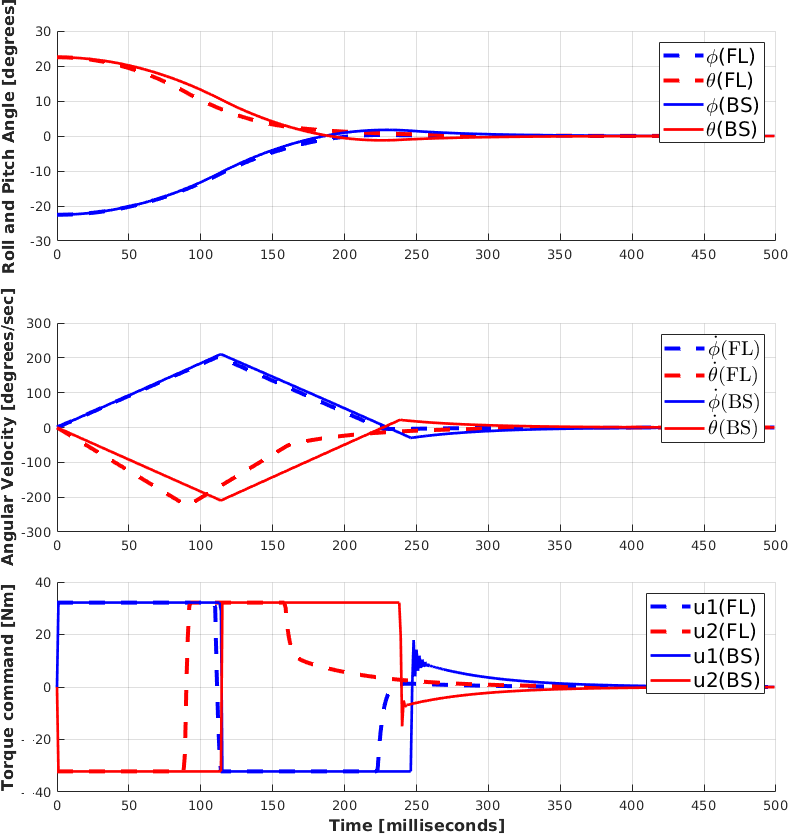}
	\caption{Simulation comparison between the Proportional-Derivative with Feedback Linearization (FL) controller and the Backstepping (BS) Controller.}
	\label{Sim1}
\end{figure}
By canceling nonlinear terms in the system dynamics using feedback linearization approach, the proposed controller can achieve all desired attitudes successfully. In this simulation, initial parameters are set to:
\begin{equation}
\begin{split}
\delta_1 = 45~ deg, \quad \delta_2 = -45~ deg,\\
    J_{Bii} = [0.662, 0.940, 1.448] \ kg \cdot m^2,\\
    J_{Wii} = [0.006565, 0.011689, 0.006565] \ kg \cdot m^2,\\
    J_{m_{W}ii} = [0.3055, 0.4103, 0.7158] \ kg \cdot m^2,\\
    \textbf{x}(0) = [-22.5, 22.5, 0]^T \ deg.
\end{split}
\end{equation}

The AGRO system has initial pitch and roll angles as -22.5 and 22.5 degrees, respectively, and the desired attitude is set as $\textbf{x}_d = [0, 0, 0]^T$ degrees. Also, the control input limit is set as $\pm$32.1521 $Nm$. The control gain values are obtained using the Linear Quadratic Regulator (LQR) method:

\begin{equation}
J(u) = \int_{0}^{inf} (x^TQx + u^TRu) dt,
\end{equation}
where $Q=diag[16.5~~ 5299.4~~ 16.5~~ 5299.4~~ 16.5~~  16.5]$, $R=diag[0.0002487~~ 0.0001326~~ 0.00006407]$ which yields

\begin{equation}
\begin{bmatrix}
k_{\varphi,1}\\
k_{\varphi,2}
\end{bmatrix}
=
\begin{bmatrix}
k_{\theta,1}\\
k_{\theta,2}
\end{bmatrix}
=
\begin{bmatrix}
k_{\psi,1}\\
k_{\psi,2}
\end{bmatrix}
=
\begin{bmatrix}
19.9977\\
122.6497
\end{bmatrix}
\end{equation}

The Backstepping controller for this comparison used $K1 = 20I$, $K2 = 1800I$, $\Lambda=I$, $\Gamma=I$ and $\Sigma=I$. 
The result of comparing the PD+FL controller and the Backstepping Controller are shown in Fig. \ref{Sim1}. Each proposed controller improves upon the simple PD controller, with both the PD+FL controller and the Backstepping controller achieving the desired attitude in about 250 milliseconds. For reference, the robot prototype takes about 400ms to fall from a 0.85 meter drop.

\subsection{Adaptive Backstepping Control}
\begin{figure}[t]
	\centering
	\includegraphics[width=1\linewidth]{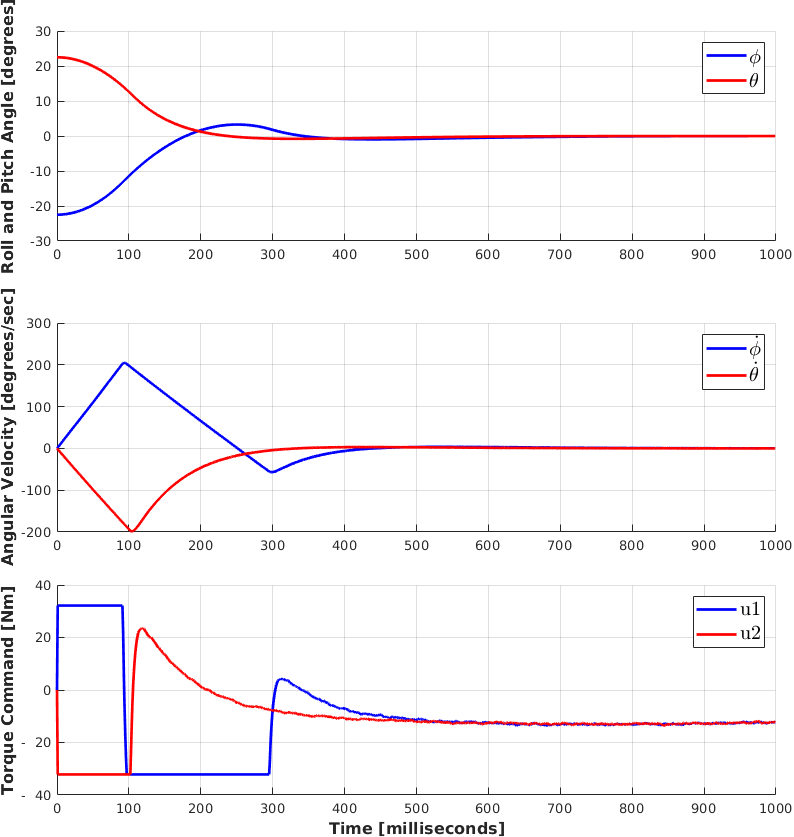}
	\caption{Adaptive Backstepping Controller rejecting disturbances in simulation.}
	\label{Sim2}
\end{figure}
The adaptation law was tested with Backstepping in the same simulation situation, but with different control parameters and added disturbances in the form of an external control input. The Backstepping controller for this simulation used $K1 = 10I$, $K2 = 200I$, $\Lambda = I$, $\Gamma = I$ and $\Sigma = 0.0005I$. The disturbance $L$ is a combination of a constant offset, a low-frequency (2 $rad/s$) sine wave, and Gaussian noise. The affect of these sources of noise and disturbance was kept under 20\%, 20\%, and 5\% of the maximum control effort, respectively.

\begin{figure}[t]
	\centering
	\includegraphics[width=1\linewidth]{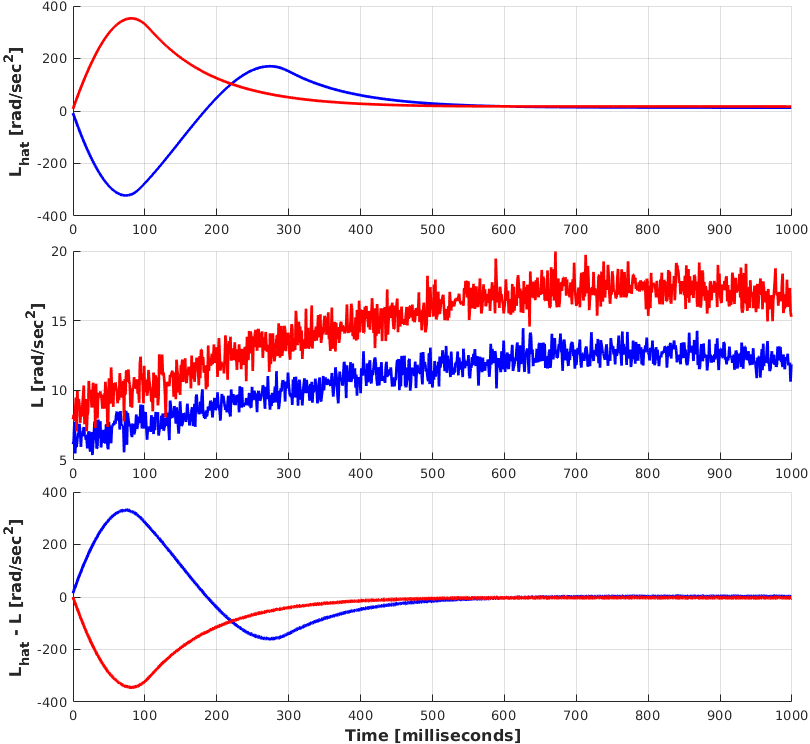}
	\caption{Adaptive Backstepping Controller rejecting disturbances in simulation. (Full time profile)}
	\label{Sim3}
\end{figure}

\begin{figure}[t]
	\centering
	\includegraphics[width=1\linewidth]{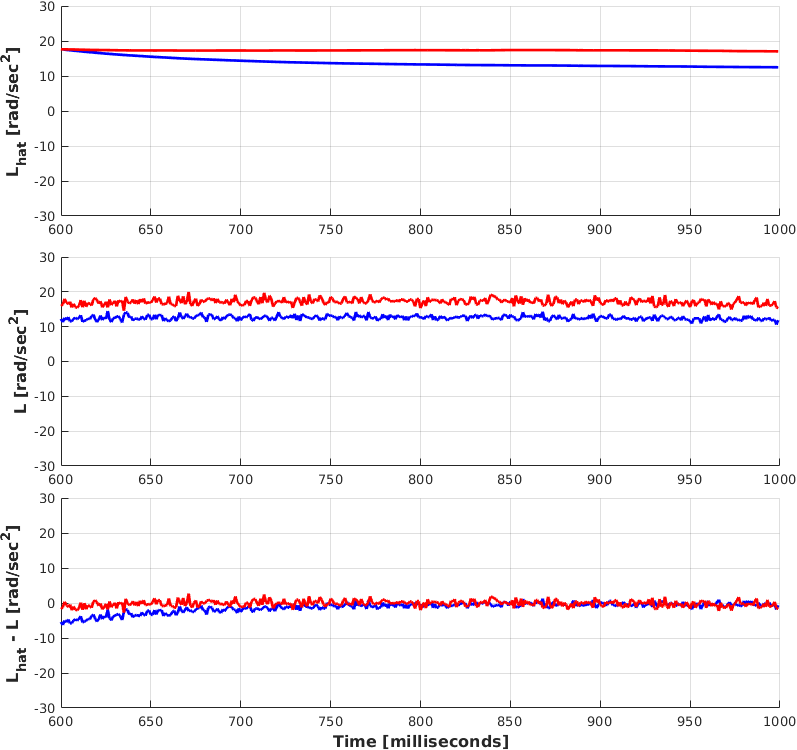}
	\caption{Adaptive Backstepping Controller rejecting disturbances in simulation. (Zoomed in from 600ms to 1000ms)}
	\label{Sim3zoom}
\end{figure}

As shown in Fig. \ref{Sim2}, the adaptation law tracks the constant offset and sinusoidal disturbance sufficiently to attain and maintain stability. Fig. \ref{Sim3} shows the external disturbance $L$, the estimate $\hat{L}$ from the adaptation law, and the error between the two over time. Transient motion of the system affects the adaptation convergence, but once the robot approaches steady-state, the adaptation law is able to compensate for the offset and 2 $rad/s$ sine wave. Fig. \ref{Sim3zoom} shows the disturbance estimate $\hat{L}$ tracking the external disturbance closely, and the error between the two approaching 0 by ~700$ms$. 



\section{Conclusion and Future Work}
The PD+FL control method was compared to the backstepping control method using the same initial conditions in simulation. Both the backstepping controller and the PD+FL controller stabilized the system within 250 milliseconds, The backstepping controller with the adaptation law also performed well at compensating for offset noisy sinusoidal disturbances.

Future work includes implementing the backstepping controller on the AGRO prototype, changing the wheel steering configuration on-the-fly to increase the control authority about the instantaneous axis of rotation, and investigating using a pre-planned trapezoidal feedforward torque profile to drive the wheels to their torque and velocity limits and optimize for minimum-time stabilization, with a feedback controller providing any additional necessary stabilization. Such a feed-forward profile, with a stabilizing controller to handle unmodeled phenomena, should give the theoretical fastest convergence possible with this system. 

\bibliographystyle{IEEEtran}
\bibliography{IEEEabrv,biblio}
\end{document}